\documentclass{article} 
\usepackage{nips13submit_e,times}
\usepackage{hyperref}
\usepackage{url}
\usepackage{amsmath}
\usepackage[]{graphicx}
\usepackage{subfigure}

\title{Gaussian-binary Restricted Boltzmann Machines on Modeling
Natural Image Statistics}

\author{\name Nan Wang \email nan.wang@ini.rub.de \\
       \name Jan Melchior \email jan.machior@ini.rub.de \\
       \name Laurenz Wiskott \email laurenz.wiskott@ini.rub.de \\
       \addr Instit\"ut f\"ur Neuroinformatik\\
       Internatial Graduate School of Neuroscience\\
       Ruhr-Universit\"at Bochum\\
       D-44780, Bochum, Germany}

\author{
Nan Wang \\
Instit\"ut f\"ur Neuroinformatik\\
Ruhr-Universit\"at Bochum\\
Bochum, 44780, Germany \\
\texttt{nan.wang@ini.rub.de} 
\And
Jan Melchior \\
Instit\"ut f\"ur Neuroinformatik\\
Ruhr-Universit\"at Bochum\\
Bochum, 44780, Germany\\
\texttt{jan.machior@ini.rub.de} 
\AND
Laurenz Wiskott\\
Instit\"ut f\"ur Neuroinformatik\\
Ruhr-Universit\"at Bochum\\
Bochum, 44780, Germany \\
\texttt{laurenz.wiskott@ini.rub.de } \\
}

\nipsfinalcopy 

\begin{document}

\maketitle

\begin{abstract}
  We present a theoretical analysis of Gaussian-binary restricted Boltzmann
  machines (GRBMs) from the perspective of density models. The key aspect of this
  analysis is to show that GRBMs can be formulated as a constrained
  mixture of Gaussians, which gives a much better insight into the model's
  capabilities and limitations. 
  We show that GRBMs are capable of learning
  meaningful
  features both in a two-dimensional blind source separation task and in
  modeling natural images. 
  Further, we show that reported difficulties in
  training GRBMs are due to the failure of the training algorithm rather than
  the model itself. Based on our analysis we are able to propose several
  training recipes, which allowed successful and fast training in our
  experiments. Finally, we discuss the relationship of GRBMs to several
  modifications that have been proposed to improve the model.
\end{abstract}

\section{Introduction}
Inspired by the hierarchical structure of the visual cortex, recent studies on
probabilistic models used deep hierarchical architectures to learn high order
statistics of the data~\cite{KarklinLewicki-2009a, KosterHyvarinen-2010a}. One
widely used architecture is a deep believe network (DBN), which is usually
trained as stacked restricted Boltzmann machines
(RBMs)~\cite{HintonSalakhutdinov-2006a, BengioLamblinEtAl-2006a,
ErhanBengioEtAl-2010a}. Since the original formulation of RBMs assumes binary
input values, the model needs to be modified in order to handle continuous
input values.  One common way is to replace the binary input units by linear
units with independent Gaussian noise, which is known as Gaussian-binary
restricted Boltzmann machines (GRBMs) or Gaussian-Bernoulli restricted
Boltzmann machines ~\cite{Krizhevsky-2009a, ChoIlinEtAl-2011b} first proposed
by \cite{WellingRosen-ZviEtAl-2004a}.

The training of GRBMs is known to be difficult, so that several modifications
have been proposed to improve the training. \cite{LeeEkanadhamEtAl-2007a} used
a sparse penalty during training, which allowed them to learn meaningful
features from natural image patches.  \cite{Krizhevsky-2009a} trained GRBMs on
natural images and concluded that the difficulties are mainly due to the
existence of high-frequency noise in the images, which further prevents the
model from learning the important structures. \cite{TheisGerwinnEtAl-2011a}
illustrated that in terms of likelihood estimation GRBMs are already
outperformed by simple mixture models. Other researchers focused on improving
the model in the view of generative
models~\cite{RanzatoKrizhevskyEtAl-2010a,RanzatoHinton-2010a,
CourvilleBergstraEtAl-2011a, LeHeessEtAl-2011a}. ~\cite{ChoIlinEtAl-2011b}
suggested that the failure of GRBMs is due to the training algorithm and
proposed some modifications to overcome the difficulties encountered in
training GRBMs.

The studies above have shown the failures of GRBMs empirically, but to our
knowledge there is no analysis of GRBMs apart from our preliminary
work~\cite{WangMelchiorEtAl-2012a}, which accounts the reasons behind these
failures. In this paper, we extend our work in which we consider GRBMs from the
perspective of density models, i.e.  how well the model learns the distribution
of the data. We show that a GRBM can be regarded as a mixture of Gaussians,
which has already been mentioned briefly in previous
studies~\cite{Bengio-2009a, TheisGerwinnEtAl-2011a,
CourvilleBergstraEtAl-2011a} but has gone unheeded. This formulation makes
clear that GRBMs are quite limited in the way they can represent data. However
we argue that this fact does not necessarily prevent the model from learning
the statistical structure in the data.  We present successful training of GRBMs
both on a two-dimensional blind source separation problem and natural image
patches, and that the results are comparable to that of independent component
analysis (ICA). Based on our analysis we propose several training recipes,
which allowed successful and fast training in our experiments.  Finally, we
discuss the relationship between GRBMs and above mentioned modifications of the
model.

\section{Gaussian-binary restricted Boltzmann machines (GRBMs)}

\subsection{The model}
A Boltzmann Machine (BM) is a Markov Random Field with stochastic
\emph{visible} and \emph{hidden} units \cite{Smolensky-1986a}, which are denoted as~\(\mathbf{X}:=\left(
  X_1,\ldots,X_M\right)^T\) and~\(\mathbf{H}:=\left( H_1, \ldots,
  H_N\right)^T \), respectively. In general, we use bold letters denote vectors and matrices.   

  The joint probability distribution is defined as
\begin{eqnarray}
    P\left( \mathbf{X},\mathbf{H} \right) &:=& \frac{1}{Z} \,
    \mathrm{e}^{-\frac{1}{T_0}E(\mathbf{X},\mathbf{H})},
    \label{eqn:jointprobOfXH}\\
    Z &:=& \int \int{\mathrm{e}^{-\frac{1}{T_0}E\left(
        \mathbf{x},\mathbf{h} \right)}} \mathrm{d} \mathbf{x}\,\mathrm{d} \mathbf{h}
    \label{eqn:partitionFunc}
\end{eqnarray}
where \(E\left( \mathbf{X},\mathbf{H} \right)\) denotes an \emph{energy
function} as known from statistical physics, which defines the dependence
between \(\mathbf{X}\) and \(\mathbf{H}\). The temperature parameter \(T_0\) is
usually ignored by setting its value to one, but it can play an important role
in inference of BMs \cite{DesjardinsCourvilleEtAl-2010a}.  The
\emph{partition function} $Z$ normalizes the probability distribution by
integrating over all possible values of \(\mathbf{X}\) and \(\mathbf{H}\),
which is intractable in most cases.  So that in training BMs using gradient
descent the partition function is usually estimated using sampling methods.
However, even sampling in BMs remains difficult due to the dependencies between
all variables.

An RBM is a special case of a BM where the
energy function contains no terms combining two different hidden or
two different visible units. Viewed as a graphical model, there are
no lateral connections within the visible or hidden layer, which results
in a bipartite graph. This implies that the hidden units are conditionally independent given the visibles and vice versa, which allows efficient sampling.
 
The values of the visible and hidden units are usually assumed to be binary,
i.e. \(X_m, H_n\in\left\{ 0, 1 \right\}\). The most common way to extend an RBM
to continuous data is a GRBM, which assumes continuous values for the visible
units and binary values for the hidden units. Its energy function
\cite{ChoIlinEtAl-2011b, WangMelchiorEtAl-2012a} is defined as
\begin{eqnarray}
  E\left( \mathbf{X}, \mathbf{H} \right) :
  &=&
     \sum_i^M \frac{\left( X_i-b_i \right)^2}{2\sigma^2}
    - \sum_j^N c_j H_j
    - \sum_{i,j}^{M,N} \frac{
         X_i w_{ij} H_j
    }{\sigma^2} 
    \\ 
  \label{eqn:energyFuncGauss1}
  &=&
    \frac{|| \mathbf{X} - \mathbf{b} ||^2}{2\sigma^2} 
    - \mathbf{c}^T \mathbf{H}
    - \frac{
        \mathbf{X}^T \mathbf{W} \mathbf{H}
    }{\sigma^2}
    ,
  \label{eqn:energyFuncGauss2}
\end{eqnarray}
where \(||\mathbf{u}||\) denotes the Euclidean norm of $\mathbf{u}$. In GRBMs
the visible units given the hidden values are Gaussian distributed with
standard deviation $\sigma$. Notice that some authors
\cite{Krizhevsky-2009a, ChoIlinEtAl-2011b, Melchior-2012} use an
independent standard deviation for each visible unit, which comes into account
if the data is not whitened \cite{Melchior-2012}.  \\ 

The conditional probability distribution of the visible given the hidden units
is given by
  \begin{eqnarray}
  P\left( \mathbf{X}|\mathbf{h} \right) 
  &=& 
      \frac{P\left( \mathbf{X}, \mathbf{h} \right)}
          {\int\limits{P\left( \mathbf{x}, \mathbf{h} \right) \mathrm{d} \mathbf{x} }}  \\
  & \stackrel{(\ref{eqn:jointprobOfXH}, \ref{eqn:energyFuncGauss2})}{=} &
      \frac{
      \mathrm{e}^{\mathbf{c}^T \mathbf{h}} 
       \prod\limits_i^M{ \mathrm{e}^{  
                         \frac{X_i \mathbf{w}_{i *}^T \mathbf{h} }{\sigma^2} 
                        - \frac{|| X_i - b_i ||^2}{2 \sigma^2}
                        } } }
        {  \int\limits{\mathrm{e}^{\mathbf{c}^T \mathbf{h}}
          \prod\limits_i^M{ \mathrm{e}^{
                         \frac{x_i \mathbf{w}_{i *}^T \mathbf{h} }{\sigma^2}
                        - \frac{|| x_i - b_i ||^2}{2 \sigma^2} }
                        \mathrm{d} \mathbf{x} } } } \\
  &=&
      \prod_i^M{
          \frac{ \mathrm{e}^{
                               \frac{X_i \mathbf{w}_{i *}^T \mathbf{h} }{\sigma^2}
                              - \frac{|| X_i - b_i ||^2}{2 \sigma^2} 
                  }
          }{  \int\limits \mathrm{e}^{
                               \frac{x_i \mathbf{w}_{i *}^T \mathbf{h} }{\sigma^2}
                              - \frac{|| x_i - b_i ||^2}{2 \sigma^2} 
                          }\mathrm{d}x_i
          }
      } 
      \label{eqn:probOfXH1}
      \\
  &\stackrel{(\ref{eqn:quadraticExp})}{=}&
      \prod_i^M{
          \frac{ \mathrm{e}^{
                              - \frac{|| X_i - b_i - \mathbf{w}_{i *}^T \mathbf{h} ||^2}{2 \sigma^2}
                  }
          }{  \int\limits \mathrm{e}^{
                              - \frac{|| x_i - b_i - \mathbf{w}_{i *}^T \mathbf{h} ||^2}{2 \sigma^2}
                          }\mathrm{d}x_i
          }
      } 
      \label{eqn:probOfXH2}
      \\
  &=&
      \prod_i^M{
          \underbrace{
          \mathcal{N}\left( X_i;b_i + \mathbf{w}_{i *}^T \mathbf{h}, \sigma^2 \right)
          }_{=\,P(X_i|\mathbf{h})}
      }
      \label{eqn:probOfXHGauss} \\
  &=&
        \mathcal{N}\left( \mathbf{X};\mathbf{b} + \mathbf{W} \mathbf{h}, \sigma^2 \right),
        \label{eqn:probOfXHGaussVector}
\end{eqnarray}

where \(\mathbf{w}_{i*}\) and \(\mathbf{w}_{* j}\) denote the \(i\)th
row and the \(j\)th column of the weight matrix, respectively.  \(
\mathcal{N}\left( x ; \mathbf{\mu} , \sigma^2 \right) \) denotes a
Gaussian distribution with mean \(\mu\) and variance \(\sigma^2\).
And \(\mathcal{N}\left( \mathbf{X};\boldsymbol\mu, \sigma^2
\right)\) denotes an isotropic multivariate Gaussian
distribution centered at vector \(\boldsymbol\mu\) with variance \(\sigma^2\) in all directions. 
From (\ref{eqn:probOfXH1}) to (\ref{eqn:probOfXH2}) we used the relation
\begin{eqnarray}
  \frac{ax}{\sigma^2}-\frac{\left( x-b \right)^2}{2\sigma^2}
  &=&
  \frac{-x^2+2bx+2ax-b^2}{2\sigma^2} \nonumber\\
  &=&
  \frac{-x^2+2bx+2ax-b^2+a^2-a^2+2ab-2ab}{2\sigma^2} \nonumber\\
  &=&
  \frac{-(x-a-b)^2+a^2+2ab}{2\sigma^2}. 
  \label{eqn:quadraticExp}
\end{eqnarray}
The conditional probability distribution of the hidden units given the visibles can be derived as follows
\begin{eqnarray}
  P(\mathbf{H}|\mathbf{x}) 
  &=& \frac{P(\mathbf{x},\mathbf{H})}{\sum\limits_{\mathbf{h}} P(\mathbf{x}, \mathbf{h})} \\ 
  &\stackrel{(\ref{eqn:jointprobOfXH}, \ref{eqn:energyFuncGauss2})}{=}& \frac{
  \mathrm{e}^{- \frac{||\mathbf{x} - \mathbf{b}||^2}{2 \sigma^2}}
      \prod\limits_j^N{ \mathrm{e}^{ \left( c_j + \frac{\mathbf{x}^T \mathbf{w}_{* j}}{\sigma^2} \right)H_j } }
  }{\sum\limits_{\mathbf{h}}{
  \mathrm{e}^{- \frac{||\mathbf{x} - \mathbf{b}||^2}{2 \sigma^2}}
          \prod\limits_j^N{ \mathrm{e}^{
                  \left( c_j + \frac{\mathbf{x}^T \mathbf{w}_{* j}}{\sigma^2} \right)h_j } } } } \\
  &=&
  \prod_j^N{
    \underbrace{
      \frac{\mathrm{e}^{
          \left( c_j + \frac{\mathbf{x}^T \mathbf{w}_{* j}}{\sigma^2} \right)H_j }
      }{\sum\limits_{h_j}{ \mathrm{e}^{
            \left( c_j + \frac{\mathbf{x}^T \mathbf{w}_{* j}}{\sigma^2} \right)h_j } } }
    }_{=\,P(H_j|\mathbf{x})}
  }. \\
  \Longrightarrow \quad P(H_j=1|\mathbf{x}) &=&
\frac{1}{1+\mathrm{e}^{-\left(c_j + \frac{\mathbf{x}^T \mathbf{w}_{* j}}{\sigma^2}\right) }}  
  \label{eqn:probOfHXGauss}
\end{eqnarray}
$P(\mathbf{H}|\mathbf{x})$ turns out to be a product of independent sigmoid functions, which is a frequently used non-linear activation function in artificial neural networks.

\subsection{Maximium likelihood estimation}

\label{sec:trainingAlgorithm}
Maximum likelihood estimation (MLE) is a frequently used technique for training
probabilistic models like BMs. In MLE we have a data set
\(\mathcal{\tilde{X}}=\left\{ \tilde{\mathbf{x}}_1, \ldots,
\tilde{\mathbf{x}}_L \right\}\) where the observations \( \tilde{\mathbf{x}}_l
\) are assumed to be independent and identically distributed (i.i.d.). The goal
is to find the optimal parameters \(\tilde{\boldsymbol\Theta}\) that maximize
the likelihood of the data, i.e. maximize the probability that the data is
generated by the model~\cite{Bishop-2006a}.  For practical reasons one often
considers the logarithm of the likelihood, which has the same maximum as the
likelihood since it is a monotonic function. The log-likelihood is defined as
\begin{equation}
  \ln P( \mathcal{\tilde{X}}; \boldsymbol\Theta ) = 
  \ln{\prod_{l=1}^L{P\left( \tilde{\mathbf{x}}_l; \boldsymbol\Theta \right)}} =
  \sum_{l=1}^L{\ln{P\left( \tilde{\mathbf{x}}_l; \boldsymbol\Theta \right)}}.
  \label{}
\end{equation}
We use the average log-likelihood per training case denoted by $\hat\ell$. For
RBMs it is defined as
\begin{equation}
    \hat\ell :=
    \left<  \ln P( \mathcal{\tilde{X}}; \boldsymbol\Theta )
	\right>_{\tilde{\mathbf{x}}} = 
	\left< 
		\ln \left( \sum_{\mathbf{h}} { 
	    \mathrm{e}^{ -E \left( \tilde{\mathbf{x}},\mathbf{h} \right) } } 
		\right)
	\right>_{\tilde{\mathbf{x}}} - \ln Z,
	\label{eqn:loglikelihood}
\end{equation}
where \(\tilde{\mathbf{x}} \in \mathcal{\tilde{X}}\). And \(\left< f(u) \right>_u\) denotes the expectation of the function
\(f(u)\) with respect to variable \(u\).

The gradient of the $\hat\ell$ turns out 
to be the difference between the expectations of the energies gradient under the data and model distribution, which is given by
  \begin{eqnarray}
	  \frac{ \partial \hat\ell}{ \partial \theta } 
      &\stackrel{(\ref{eqn:loglikelihood}, \ref{eqn:partitionFunc})}{=}& 
	\left<
	    \sum\limits_{\mathbf{h}}{
	    \frac{ \frac{\mathrm{e}^{-E\left( \tilde{\mathbf{x}},\mathbf{h} \right)}}{Z}
	    }{ \sum\limits_{\mathbf{h'}}{ \frac{\mathrm{e}^{-E \left( \tilde{\mathbf{x}},\mathbf{h'} \right)}}{Z} } }
		\left( - \frac{ \partial E \left( \tilde{\mathbf{x}}, \mathbf{h} \right) }{ \partial \theta } \right)
	    }
	\right>_{\tilde{\mathbf{x}}} -
	\frac{1}{Z} \sum\limits_{\mathbf{h}} \sum\limits_{\mathbf{x}} { \mathrm{e}^{ -E \left( \mathbf{x}, \mathbf{h} \right) 
	    }\left(  
		- \frac{ \partial E \left( \mathbf{x},\mathbf{h} \right) }{ \partial \theta }
	    \right)
	} \\
	&\stackrel{(\ref{eqn:jointprobOfXH})}{=}& 
		- \left<
			\sum_{\mathbf{h}}{
				P\left( \mathbf{h}|\tilde{\mathbf{x}} \right)
				\frac{ \partial E \left( \tilde{\mathbf{x}},\mathbf{h} \right) }{ \partial \theta }
			}
		\right>_{\tilde{\mathbf{x}}} +
		\left<
			\sum_{\mathbf{h}}{
				P\left( \mathbf{h}|\mathbf{x} \right)
				\frac{ \partial E \left( \mathbf{x},\mathbf{h} \right) }{ \partial \theta }
			}
		\right>_{\mathbf{x}}.
	\label{eqn:loglikelihooddiff}
  \end{eqnarray}
In practice, a finite set of i.i.d. samples can be used to approximate the
expectations in (\ref{eqn:loglikelihooddiff}).  While we can use the training
data to estimate the first term, we do not have any i.i.d. samples from the
unknown model distribution to estimate the second term.  Since we are able to
compute the conditional probabilities in RBMs efficiently, Gibbs sampling can
be used to generate those samples. But Gibbs-sampling only guarantees to
generate samples from the model distribution if we run it infinite long. As
this is impossible, a finite number of \(k\) sampling steps are used instead.
This procedure is known as Contrastive Divergence~-~\(k\) (CD-\(k\)) algorithm,
in which even \(k=1\) shows good results~\cite{Hinton-2002a}. The CD-gradient
approximation is given by
  \begin{equation}
    \frac{
    \partial \hat\ell
    }{
	\partial \theta
    } \approx
    - \left<
	  \sum_{\mathbf{h}}{
		  P\left( \mathbf{h}|\tilde{\mathbf{x}}  \right)
	       \frac{
		  \partial E \left( 
		      \tilde{\mathbf{x}} ,\mathbf{h}
		  \right)
	      }{
		  \partial \theta
	      }
	  }
      \right>_{ \tilde{\mathbf{x}} } +
      \left<
	  \sum_{\mathbf{h}}{
		  P( \mathbf{h}|\mathbf{x}^k )
	       \frac{
		  \partial E \left( 
		      \mathbf{x}^{(k)},\mathbf{h}
		  \right)
	      }{
		  \partial \theta
	      }
	  }
      \right>_{\mathbf{x}^{(k)}},
      \\
	\label{eqn:loglikelihooddiffapprox1}
  \end{equation}
where \(\mathbf{x}^{(k)}\) denotes the samples after \(k\) steps of Gibbs sampling.
The derivatives of the GRBM's energy function with respect to the parameters are given by
  \begin{eqnarray}
	\frac{
		\partial E\left(
				\mathbf{X}, \mathbf{H}
			\right)
	}{
		\partial \mathbf{b}
	}
	&=&
	- \frac{\mathbf{X} - \mathbf{b}}{\sigma^2}, \\
	\label{eqn:energydiffvisbiasGauss}
	\frac{
		\partial E\left(
				\mathbf{X}, \mathbf{H}
			\right)
	}{
		\partial \mathbf{c}
	} 
	&=&
	- \mathbf{H}, \\
	\label{eqn:energydiffhidbiasGauss}
	\frac{
		\partial E\left(
				\mathbf{X}, \mathbf{H}
			\right)
	}{
		\partial \mathbf{W}
	} 
	&=&
	- \frac{
		\mathbf{X} \mathbf{H}^T
	}{\sigma^2}, 
	\label{eqn:energydiffweightGauss}\\
	\frac{ \partial E \left( \mathbf{X},\mathbf{H} \right)
	}{ \partial\,\mathbf{\sigma} }
	&=&
	- \frac{ || \mathbf{X} - \mathbf{b} ||^2 }{ \sigma^3 }
	+ \frac{ 2 \, \mathbf{X}^T\mathbf{W}\mathbf{H} }{ \sigma^3 },
	\label{eqn:energydiffvarianceGauss}
\end{eqnarray}
and the corresponding gradient approximations~(\ref{eqn:loglikelihooddiffapprox1}) become
  \begin{eqnarray}
	\frac{ \partial \hat\ell }{ \partial \mathbf{b} }
	& \approx &
    \left<\frac{ \tilde{\mathbf{x}} -\mathbf{b}
    }{\sigma^2}\right>_{\tilde{\mathbf{x}}} 
    - \left<\frac{ \mathbf{x}^{(k)} -\mathbf{b} }{\sigma^2}\right>_{\mathbf{x}^{(k)}}
    ,
	\label{eqn:loglikelihooddiffvisbiasGauss}\\
	\frac{ \partial \hat\ell }{ \partial \mathbf{c} }
	& \approx &
	\left< P\left( \mathbf{h} = \mathbf{1}|\tilde{\mathbf{x}} \right) \right>_{\tilde{\mathbf{x}}} 
    - \left< P\left( \mathbf{h} = \mathbf{1}|\mathbf{x}^{(k)} \right) \right>_{\mathbf{x}^{(k)}}
    ,
	\label{eqn:loglikelihooddiffhidbiasGauss}\\
	\frac{
		\partial \hat\ell
	}{
		\partial \mathbf{w}
	}
	& \approx &
	\left<
		\frac{\tilde{\mathbf{x}}P\left( 
			\mathbf{h} = \mathbf{1}|\tilde{\mathbf{x}}
		\right)^T}{\sigma^2}
	\right>_{\tilde{\mathbf{x}}} - 
	\left<
	\frac{\mathbf{x}^{(k)}P\left( 
	    \mathbf{h} = \mathbf{1}|\mathbf{x}^{(k)}
	\right)^T}{\sigma^2}
    \right>_{\mathbf{x}^{(k)}}
    ,
	\label{eqn:loglikelihooddiffweightGauss}\\
  \frac{
		\partial \hat\ell
	}{
		\partial\,\mathbf{\sigma}
	}
	& \approx &
	\left<
		     \frac{
			|| \tilde{\mathbf{x}} - \mathbf{b} ||^2
		-
		2  \, \tilde{\mathbf{x}}^T\mathbf{W}\, P\left( 
			\mathbf{h} = \mathbf{1}|\tilde{\mathbf{x}}
		\right) }{\sigma^3}
  \right>_{ \tilde{\mathbf{x}} } \\
  \nonumber
  &&-\left<
		     \frac{
	    || \mathbf{x}^{(k)} - \mathbf{b} ||^2
		-
	2  \, {\mathbf{x}^{(k)}}^T\mathbf{W}\, P\left( 
	    \mathbf{h} = \mathbf{1}|\mathbf{x}^{(k)}
		\right) }{\sigma^3}
    \right>_{ \mathbf{x}^{(k)} },
  \label{eqn:loglikelihooddiffvarianceGauss}
  \end{eqnarray}
where \(P\left( \mathbf{h} = 1|\mathbf{x} \right) := (P\left( h_1 = 1 |
\mathbf{x} \right), \cdots, P\left( h_N = 1 | \mathbf{x} \right))^T\), i.e.
\(P\left( \mathbf{h} = 1|\mathbf{x} \right)\) denotes a vector of probabilities. 

\subsection{The marginal probability distribution of the visible units}

From the perspective of density estimation, the performance
of the model can be assessed by examining how well the model estimates the data
distribution. We therefore take a look at the model's marginal probability
distribution of the visible units, which can be formalized as a product of
experts (PoE) or as a mixture of Gaussians (MoG))\footnote{Some
	part of this analysis has been previously reported
	by~\cite{FreundHaussler-1992a}. Thanks to the anonymous reviewer for
	pointing out this coincidence.}.

\subsubsection{In the Form of Product of Experts}
We derive the marginal probability distribution of the visible units
$P(\mathbf{X})$ by factorizing the joint probability distribution over the
hidden units.
\begin{eqnarray}
	P\left( \mathbf{X} \right) 
    &=& 
    \sum\limits_{
		\mathbf{h}
	}{ P\left( 
	    \mathbf{X},\mathbf{h} 
		\right)
	} \\
    & \stackrel{(\ref{eqn:jointprobOfXH},\ref{eqn:energyFuncGauss1})}{=} &
	\frac{1}{Z} \,\mathrm{e}^{ 
	- \frac{|| \mathbf{X}-\mathbf{b} ||^2}{2\sigma^2} 
    }
    \prod_j^N{\sum_{h_j}{
    \mathrm{e}^{ c_j+\frac{\mathbf{X}^T\mathbf{w}_{* j} }{\sigma^2} h_j } } }
	\\
    &\stackrel{h_j\in \left\{ 0,1 \right\}}{=}&	\frac{1}{Z} 
	\prod_j^N{\left( \,\mathrm{e}^{ - \frac{|| \mathbf{X}-\mathbf{b} ||^2}{2N\sigma^2}} 
	  + \mathrm{e}^{ c_j+\frac{\mathbf{X}^T\mathbf{w}_{* j} }{\sigma^2} 
	  - \frac{|| \mathbf{X}-\mathbf{b} ||^2}{2N\sigma^2}} \right)} 
	\label{eqn:probOfXGauss1} \\
    &\stackrel{(\ref{eqn:quadraticExp})}{=}& \frac{1}{Z}
    \prod_j^N \left( \,\mathrm{e}^{ - \frac{|| \mathbf{X}-\mathbf{b} ||^2}{2N\sigma^2}}
    + \mathrm{e}^{\frac{||\mathbf{b}+N \mathbf{w}_{* j}||^2 
	      - ||\mathbf{b}||^2}{2N\sigma^2} 
	      + c_j - \frac{|| \mathbf{X}-\mathbf{b}-N \mathbf{w}_{* j} ||^2}{2N\sigma^2}}\right) \\
    &=& \frac{1}{Z} \prod_j^N{\left( \sqrt{2\pi N \sigma^2} \right)^{M}} 
	  \Big[ 
	    \mathcal{N}\left( \mathbf{X};\mathbf{b}, N\sigma^2 \right) \nonumber  \\
	 && 
	    +\,\mathrm{e}^{\frac{||\mathbf{b}+N \mathbf{w}_{* j}||^2 
	      - ||\mathbf{b}||^2}{2N\sigma^2} 
	      + c_j}
	      \mathcal{N}\left( \mathbf{X};\mathbf{b}+N\mathbf{w}_{* j}, N\sigma^2 \right)
	  \Big]
	\\
    &=:&\frac{1}{Z} \prod_j^N{p_j\left( \mathbf{X} \right)}
    \label{eqn:probOfXGauss2}.
      \end{eqnarray}

\begin{figure}[ht]
  \begin{center}
    \includegraphics[scale=0.7]{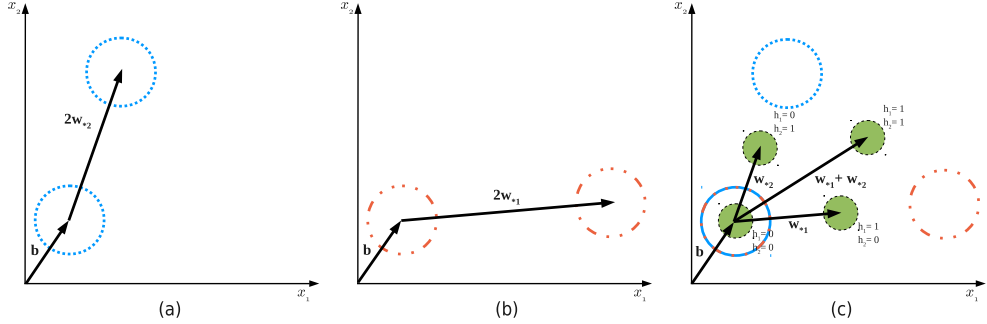}
  \end{center}
  \caption{Illustration of a GRBM-2-2 as a PoE and MoG, in which arrows indicate the roles of the
    visible bias vector and the weight vectors. (a)~and (b) visualize
    the two experts of the GRBM. The red (dotted) and blue (dashed) circles indicate the
    center of the two Gaussians in each expert. (c)~visualizes the
    components in the GRBM.  Denoted by the green (filled) circles, the four
    components are the results of the product of the two experts.  Notice how each component sits
    right between a red (dotted) and a blue (dashed) circle.}
  \label{fig:GRBMPdf}
\end{figure}

Equation~(\ref{eqn:probOfXGauss2}) illustrates that \(P(\mathbf{X})\)
can be written as a product of \(N\) factors, referred to as a
product of experts~\cite{Hinton-2002a}. Each expert
$p_j(\mathbf{X})$ consists of two isotropic Gaussians with the same variance
\(N\sigma^2\).  The first Gaussian is placed at the visible bias
\(\mathbf{b}\). The second Gaussian is shifted relative to the first one by
\(N\) times the weight vector \(\mathbf{w}_{* j}\) and scaled by a factor that
depends on \(\mathbf{w}_{* j}\) and \(\mathbf{b}\).  Every hidden
unit leads to one expert, each mode of which corresponds to one state
of the corresponding hidden unit. 
Figure~\ref{fig:GRBMPdf} (a) and (b) illustrate \(P(\mathbf{X})\) of a
GRBM-2-2 viewed as a PoE, where GRBM-$M$-$N$ denotes a GRBM with $M$ visible and $N$ hidden units.
\subsubsection{In the Form of Mixture of Gaussians}
\label{sec:lowerOrderComp}
Using Bayes'theorem, the marginal probaility of $\mathbf{X}$ can also be formalized as: 
\begin{eqnarray}
	P\left( \mathbf{X} \right) 
    &=& 
    \sum\limits_{ \mathbf{h} }{ P\left( \mathbf{X}|\mathbf{h} \right) P(\mathbf{h}) } \\
    \label{eqn:probOfXComponent}
    & \stackrel{}{=} &
    \sum\limits_{\mathbf{h}}{\mathcal{N}\left( \mathbf{X};\mathbf{b}+\mathbf{W}\mathbf{h}, \sigma^2 \right) 
    \frac{\left( \sqrt{2\pi\sigma^2} \right)^M}{Z}\,
    \mathrm{e}^{\mathbf{c}^T \mathbf{h} + \frac{|| \mathbf{b} + \mathbf{W} \mathbf{h}||^2 - ||\mathbf{b}||^2}{2\sigma^2}} } 
    \label{eqn:probOfXComponent1}\\
    & \stackrel{}{=} &
    \underbrace{  
     \frac{\left( \sqrt{2\pi\sigma^2} \right)^{M}}{Z}
	  }_{P(\mathbf{h} : \mathbf{h} \in \mathcal{H}_0)}\, \mathcal{N}\left(
	   \mathbf{X};\mathbf{b},\sigma^2 \right) 
	\nonumber \\
	  && + \sum_{j=1}^N{
		\underbrace{           
	  \frac{\left( \sqrt{2\pi\sigma^2} \right)^{M}}{Z} \,          
	  \mathrm{e}^{\frac{||\mathbf{b}+\mathbf{w}_{* j}||^2 - ||\mathbf{b}||^2}{2\sigma^2} + c_j} 
	  }_{P(\mathbf{h}_j : \mathbf{h}_j \in \mathcal{H}_1)}
	  \mathcal{N}\left( \mathbf{X};\mathbf{b}+\mathbf{w}_{* j}, \sigma^2 \right)}  
    \nonumber \\
	  && + \sum_{j=1}^{N-1}{ \sum_{k>j}^N{
	  \underbrace{  
	      \frac{\left( \sqrt{2\pi\sigma^2} \right)^{M}}{Z}             
	      \, \mathrm{e}^{\frac{||\mathbf{b}+\mathbf{w}_{* j}+\mathbf{w}_{* k}||^2 - ||\mathbf{b}||^2}{2\sigma^2} + c_j + c_k}
	      }_{P(\mathbf{h}_{jk} : \mathbf{h}_{jk} \in \mathcal{H}_2)} 
	  \mathcal{N}\left( \mathbf{X};\mathbf{b}+\mathbf{w}_{*
				      j}+\mathbf{w}_{* k}, \sigma^2 \right)
				    }
				}            
    \nonumber
    \\
	  && + \ldots , 
    \label{eqn:probOfXExpandGauss}
\end{eqnarray}
where $\mathcal{H}_k$ denotes the set of all possible binary vectors with exactly $k$ ones and $M-k$ zeros respectively. As an example, $\sum_{j=1}^{N-1}\sum_{k>j}^{N}P(\mathbf{h}_{jk}:\mathbf{h}_{jk} \in \mathcal{H}_{2}) = \sum_{\mathbf{h}\in\mathcal{H}_2}{P}(\mathbf{h})$ sums over the probabilities of all binary vectors having exactly two entries set to one. $P(\mathbf{H})$ in~(\ref{eqn:probOfXComponent1}) is derived as follows 
\begin{eqnarray}
  P(\mathbf{H})
  &=&
  \int\limits{P\left( \mathbf{x}, \mathbf{H} \right) \mathrm{d} \mathbf{x} }
  \\
  & \stackrel{(\ref{eqn:jointprobOfXH},\ref{eqn:energyFuncGauss2})}{=} &
  \frac{1}{Z} \int\limits{\mathrm{e}^{\mathbf{c}^T \mathbf{H}}
	  \prod\limits_i^M{ \mathrm{e}^{
			 \frac{x_i \mathbf{w}_{i *}^T \mathbf{H} }{\sigma^2}
			- \frac{|| x_i - b_i ||^2}{2 \sigma^2} }
			\mathrm{d} \mathbf{x} } } 
  \\
  &=&
  \frac{\mathrm{e}^{\mathbf{c}^T \mathbf{H}}}{Z} \prod\limits_i^M{
  \int\limits{ \mathrm{e}^{
			   \frac{x_i \mathbf{w}_{i *}^T \mathbf{H} }{\sigma^2}
			  - \frac{|| x_i - b_i ||^2}{2 \sigma^2} }
			  \mathrm{d} x_i } }
  \\
  &\stackrel{(\ref{eqn:quadraticExp})}{=}&
  \frac{\mathrm{e}^{\mathbf{c}^T \mathbf{H}}}{Z} \prod\limits_i^M{\left(  
  \mathrm{e}^{\frac{ (b_i+w_{i *}^T\mathbf{H})^2 - b_i^2}{2\sigma^2}}
  \int\limits{ \mathrm{e}^{ \frac{|| x_i - b_i - \mathbf{w}_{i *}^T \mathbf{H} ||^2}{2\sigma^2} }
  \mathrm{d} x_i } \right)}
  \\
  &=&
  \frac{\mathrm{e}^{\mathbf{c}^T \mathbf{H}}}{Z} \left( \sqrt{2\pi\sigma^2} \right)^M 
  \mathrm{e}^{\sum\limits_i^M{\frac{( b_i + \mathbf{w}_{i *}^T \mathbf{H})^2 - b_i^2}{2\sigma^2}} }
  \\
  &=&
  \frac{\left( \sqrt{2\pi\sigma^2} \right)^M}{Z}\,
  \mathrm{e}^{\mathbf{c}^T \mathbf{H} + \frac{|| \mathbf{b} + \mathbf{W} \mathbf{H}||^2 - ||\mathbf{b}||^2}{2\sigma^2}}
  \label{eqn:probOfHGauss}
\end{eqnarray}

Since the form in (\ref{eqn:probOfXExpandGauss}) is similar to a mixture of
isotropic Gaussians, we follow its naming convention. Each Gaussian
distribution is called a \emph{component} of the model distribution, which
is exactly the conditional probability of the visible units given a particular state of the hidden units. As well as in MoGs, each component has a \emph{mixing coefficient}, which is the marginal probability of the corresponding 
state and can also be viewed as the prior probability of 
picking the corresponding component. The total number of components in a
GRBM is $2^N$, which is exponential in the number of hidden units, see
Figure~\ref{fig:GRBMPdf}~(c) for an example. 

The locations of the components in a GRBM are not independent of each
other as it is the case in MoGs. They are centered at $\mathbf{b}+\mathbf{W}\mathbf{h}$, which is the vector sum of
the visible bias and selected weight vectors. The selection is done by the corresponding entries in $\mathbf{h}$ taking the value one.
This implies that only the $M+1$ components that sum over exactly one or zero 
weights can be placed and scaled independently. We name them first order components and the anchor component respectively. All $2^N - M - 1$ higher order components are then determined by the choice of the anchor and first order components. This indicates that GRBMs are constrained MoGs with isotropic components.

\section{Experiments}

\subsection{Two-dimensional blind source separation}
The general presumption in the analysis of natural images is that they can be
considered as a mixture of independent super-Gaussian sources
\cite{BellSejnowski-1997}, but see \cite{ZetzscheRohrbein-2001} for an analysis
of remaining dependencies.  In order to be able to visualize how GRBMs model
natural image statistics, we use a mixture of two independent Laplacian
distributions as a toy example. 

The independent sources \(\mathbf{s}=\left( s_1, s_2 \right)^T\) are
mixed by a random mixing matrix \(\mathbf{A}\) yielding
\begin{equation}
  \tilde{\mathbf{x}}' = \mathbf{A} \mathbf{s},
  \label{eqn:MixData}
\end{equation}
where $
  p\left( s_i \right) = \frac{\mathrm{e}^{-\sqrt{2}|s_i|}}{\sqrt{2}}$.
It is common to whiten the data (see Section~\ref{sec:preprocessing}), resulting in 
\begin{equation}
  \tilde{\mathbf{x}} = \mathbf{V} \tilde{\mathbf{x}}' = \mathbf{V} \mathbf{A} \mathbf{s},
  \label{eqn:MixDataWht}
\end{equation}
where \(\mathbf{V}={\left< \tilde{\mathbf{x'}}\tilde{\mathbf{x'}}^T\right>
}^ {-\frac{1}{2}}\) is the whitening matrix calculated with
principle component analysis (PCA). Through all this paper, we used
the whitened data.

In order to assess the performance of GRBMs in modeling the data distribution,
we ran the experiments for $200$ times and calculated the $\hat\ell$ for test data
analytically. For comparision, we also calculated the $\hat\ell$ over the test data for
ICA\footnote{\label{ICA_algorithm}For the fast ICA algorithm~\cite{Hyvarinen-1999b} we used for training, the $\hat\ell$ for super Gaussian sources can also be assessed analytically by 
$\hat\ell = -
\Big< \sum\limits_{j=1}^{N}\ln{2\cosh^2{\mathbf{w}_{* j}^T
\tilde{\mathbf{x}}_l}} \Big>_{\tilde{\mathbf{x}}_l} + \ln{|\det \mathbf{W}|}$.}, 
an isotropic two-dimensional Gaussian distribution and the true data
distribution\footnote{\label{ft:trueLL}As we know the true data distribution,
the exact $\hat\ell$ can be calculated by
${\hat\ell = 
- \sqrt{2}\Big< |\mathbf{u}_{1 *}\tilde{\mathbf{x}}_l| 
+ |\mathbf{u}_{2 *}\tilde{\mathbf{x}}_l| \Big>_{\tilde{\mathbf{x}}_l} 
- \ln{2}+ \ln{|\det \mathbf{U}|}}$, where $\mathbf{U} = (\mathbf{V}\mathbf{A})^{-1}$. }. 
The results are presented in Table~\ref{tab:BlindSrcSept}, which confirm the
conclusion of~\cite{TheisGerwinnEtAl-2011a} that GRBMs are not as good as ICA
in terms of $\hat\ell$.

\begin{table}[ht] 
\caption{Comparision of $\hat\ell$ between different models} 
\label{tab:BlindSrcSept} 
\begin{center} 
\begin{tabular}{ccc} 
\multicolumn{1}{c}{}  &\multicolumn{1}{c}{\bf $\hat\ell$ $\pm$ std} \\
\hline \\ 
Gaussian & $-2.8367\pm0.0086$ \\
GRBM   & $-2.8072\pm0.0088$  \\ 
ICA    & $-2.7382\pm0.0091$  \\ 
data distribution & $-2.6923\pm0.0092$ \\
\end{tabular} 
\end{center} 
\end{table} 

\begin{figure}[ht]
\centering
  \subfigure[]{
  \includegraphics[scale=0.3]{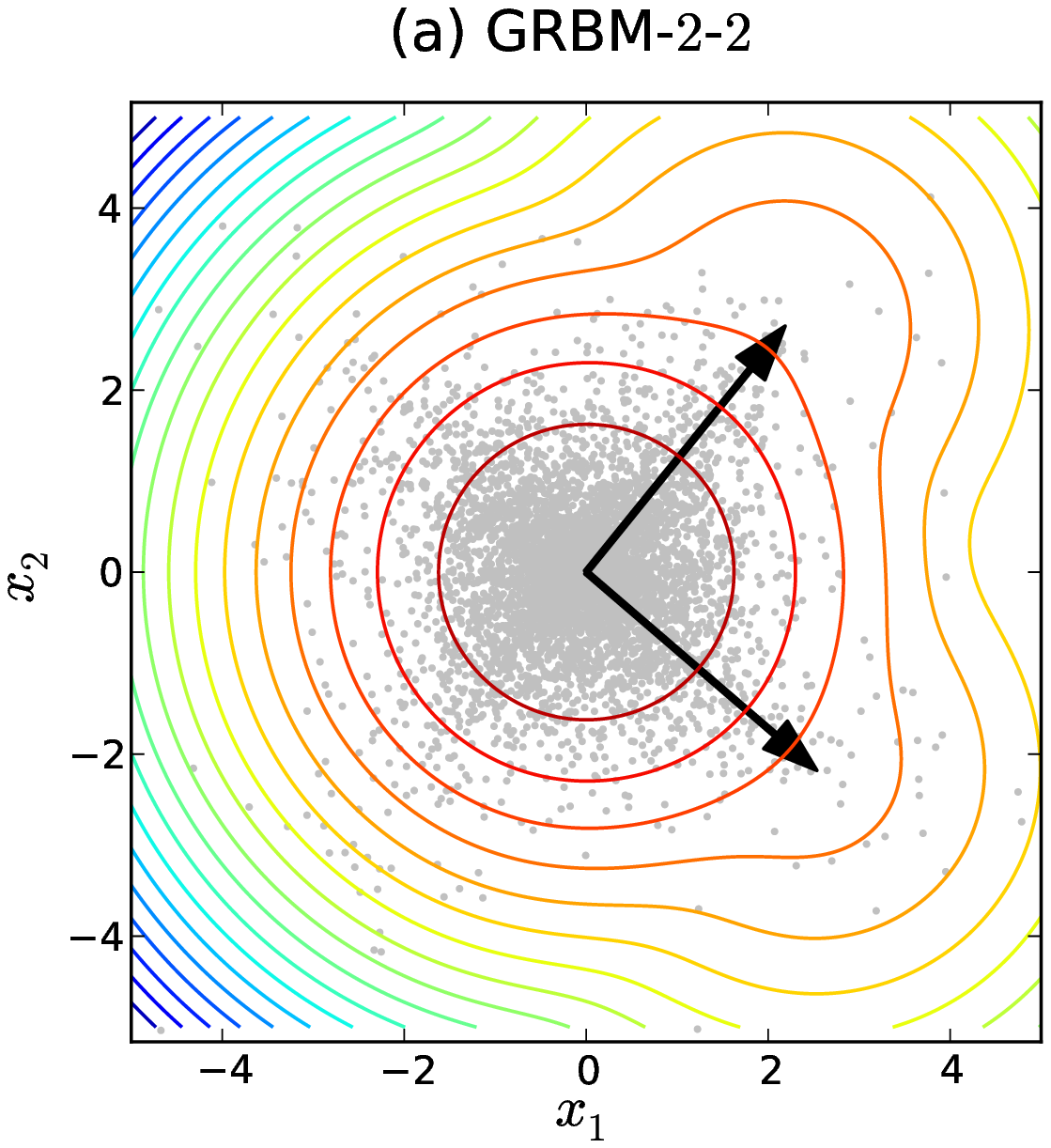}
  }
  \subfigure[]{
  \includegraphics[scale=0.3]{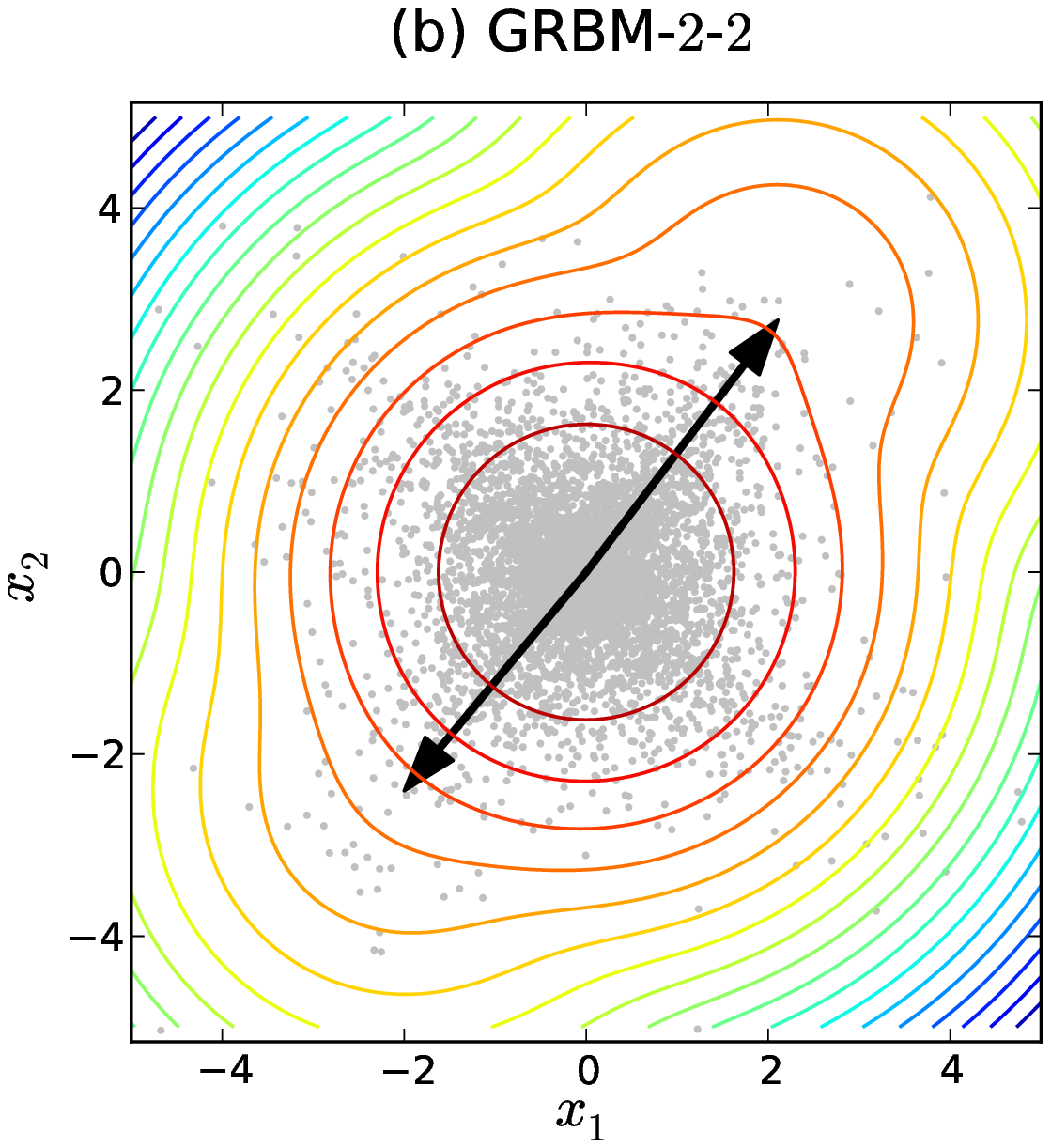}
  }
  \\
  \subfigure[]{
  \includegraphics[scale=0.3]{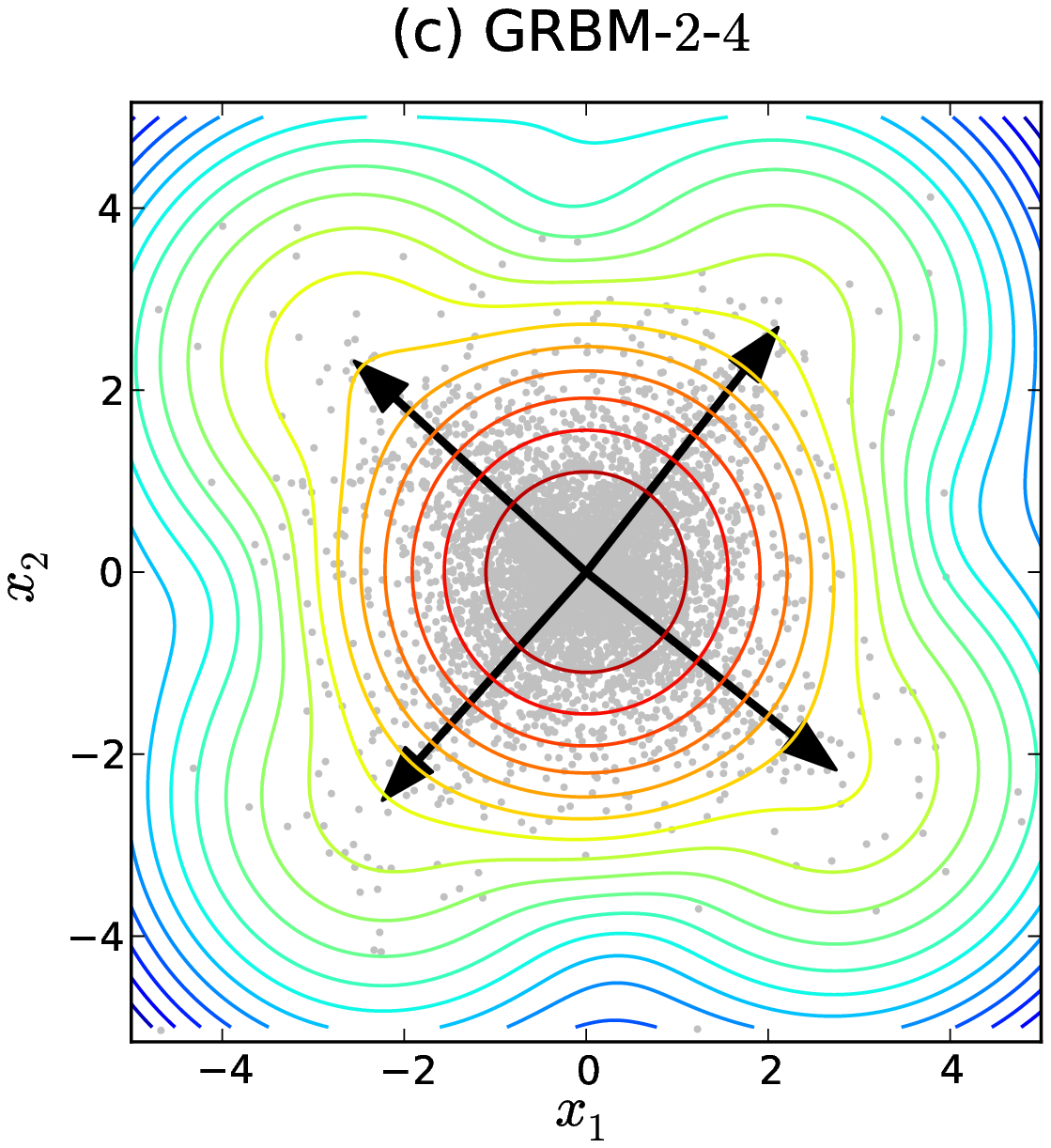}
  }
  \subfigure[]{
  \includegraphics[scale=0.3]{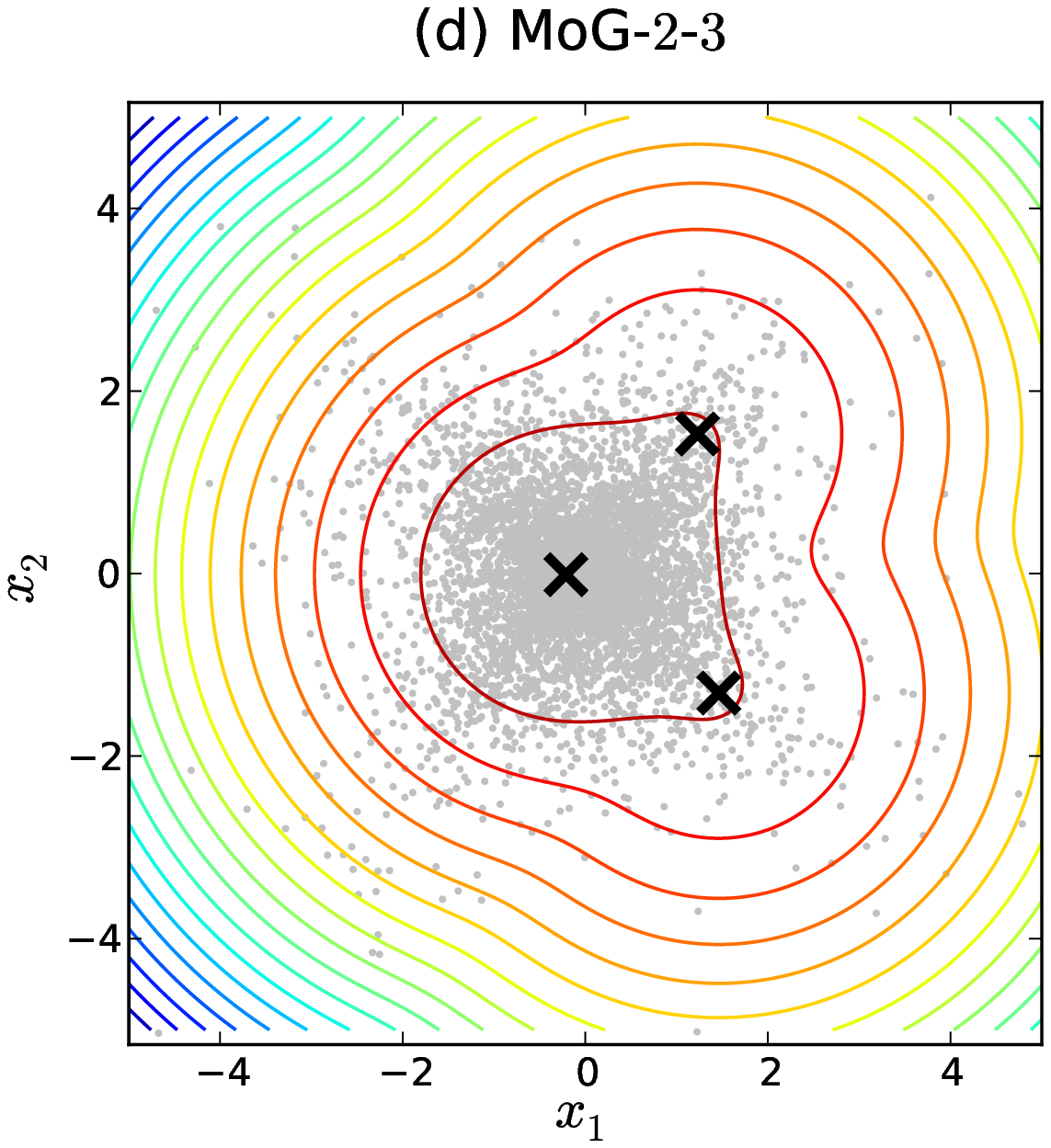}
  }
  \caption{Illustration of the log-probability densities. The data is plotted
  in blue dots. (a)GRBM-2-2 which learned two independent components.
  (b)GRBM-2-2 which learned one independent component with opposite directions. 
  (c)GRBM-2-4. 
  (d)An isotropic MoG with three components. The arrows indicate the weight vectors of GRBM, while the
  crosses denote the means of the MoG components. Comparing (a) and (d), the
  contribution of the second order component is so insignificant that the probability
  distribution of the GRBM with four components is almost the same as the MoG with only
  three components.}
  \label{fig:BlindSrcSept_GRBM}
\end{figure}

\begin{figure}[ht]
  \begin{center}
    \includegraphics[scale=0.4]{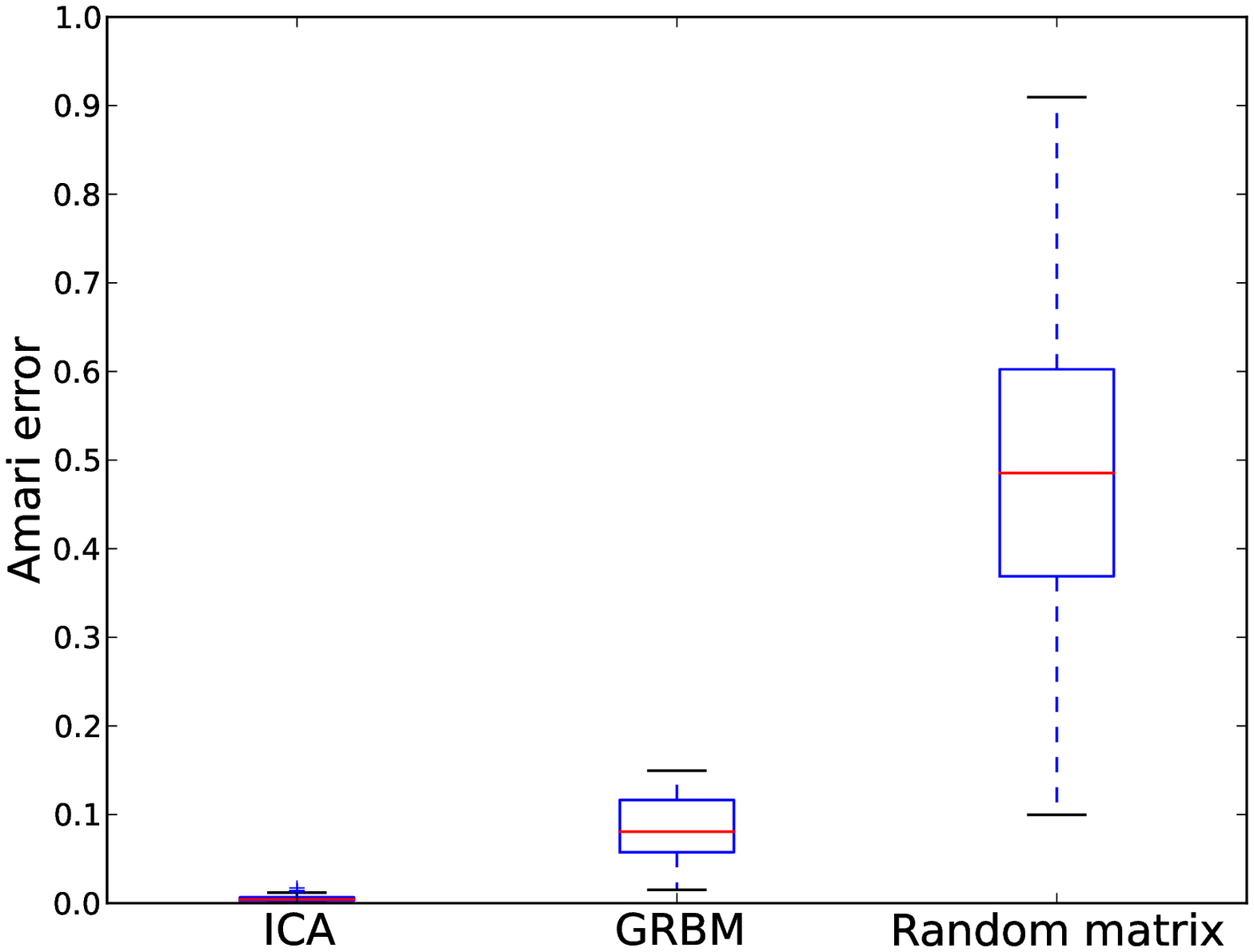}
  \end{center}
  \caption{The Amari errors between the 
  real unmixing matrix and the estimations from ICA and the 110 GRBMs. The box extends from the
  lower to the upper quantile values of the data, with a line at the median. The
  whiskers extend from the box to show the range of the reliable data points. The outlier points are
  marked by ``+''. As a base line, the amari errors between the real unmixing
  matrices and random matrices are provided.}
  \label{fig:AmariError}
\end{figure}

To illustrate how GRBMs model the statistical structure of the data, we looked at
the probability distributions of the 200 trained GRBMs. 
About half of them (110 out of 200) recovered the independent components, see Figure~\ref{fig:BlindSrcSept_GRBM} (a) as an
example. This can be further illustrated by plotting the Amari
errors\footnote{The Amari error~\cite{BachJordan-2002a} between
 two matrices $\mathbf{A}$ and $\mathbf{B}$ is defined as: $\frac{1}{2N}\left(
    \sum\limits_{i=1}^{N} \sum\limits_{j=1}^{N} {
      \frac{|(\mathbf{A}\mathbf{B}^{-1})_{ij}|}{\max_k{|(\mathbf{A}\mathbf{B}^{-1})_{ik}|}} +
      \frac{|(\mathbf{A}\mathbf{B}^{-1})_{ij}|}{\max_k{|(\mathbf{A}\mathbf{B}^{-1})_{kj}|}} }\right)- 1$.} 
between the true unmixing matrix $\mathbf{A}^{-1}$ and estimated model
matrices, i.e. the unmixing matrix of ICA and the weight matrix of the GRBM, as
shown in Figure~\ref{fig:AmariError}.  One can see that these 110 GRBMs
estimated the unmixing matrix quite well, although GRBMs are not as good as
ICA. This is due to the fact that the weight vectors in GRBMs are not
restricted to be orthogonal as in ICA. 

For the remaining 90 GRBMs, the two weight vectors pointed to the opposite
direction as shown in Figure~\ref{fig:BlindSrcSept_GRBM} (b). Accordingly,
these GRBMs failed to estimate the unmixing matrix, but in terms of density
estimation these solutions have the same quality as the orthogonal ones. Thus
all the 200 GRBMs were able to learn the statistical structures in the data and
model the data distribution pretty well. 

For comparison, we plotted the
probability distribution of a learned GRBM with four hidden units, see
Figure~\ref{fig:BlindSrcSept_GRBM} (c), in which GRBMs can always find the two
independent components correctly.

To further show how the components contribute to the model distribution, we
randomly chose one of the 110 GRBMs and
calculated the mixing coefficients of the anchor 
and the first order components 
, as shown in
Table~\ref{tab:BlindSrcSept_coefficients}. The large mixing coefficient for the anchor component indicates that the model will most likely reach hidden states in which none of the hidden units are activated. In general, the more activated hidden units a state has, the less likely it will be reached, which leads naturally to a sparse representation of the data.
\begin{table}[ht] 
  \caption{The mixing coefficients of a successfully-trained GRBM-2-2, GRBM-2-4 and an MoG-3.} 
\label{tab:BlindSrcSept_coefficients} 
\begin{center} 
\begin{tabular}{cccccc} 
\multicolumn{1}{c}{} & $\sum\limits_{\mathbf{h}\in\mathcal{H}_0}{P}(\mathbf{h})$ 
    & $\sum\limits_{\mathbf{h}\in\mathcal{H}_1}{P}(\mathbf{h})$ 
    & $\sum\limits_{\mathbf{h}\in\mathcal{H}_2}{P}(\mathbf{h})$ 
    & $\sum\limits_{\mathbf{h}\in\mathcal{H}_3}{P}(\mathbf{h})$ 
    & $\sum\limits_{\mathbf{h}\in\mathcal{H}_4}{P}(\mathbf{h})$ \\
\hline \\ 
GRBM-2-2 & $0.9811$ & $0.0188$ & $7.8856 e$-$05$ & -- & --\\ 
GRBM-2-4 & $0.9645$ & $0.0352$ & $3.4366 e$-$04$ & $1.2403 e$-$10$ & $6.9977 e$-$18$\\ 
MoG-3 & $0.9785$ & $0.0215$ & -- & -- & --\\
\end{tabular} 
\end{center} 
\end{table} 

The dominance of $\sum_{\mathbf{h}\in\mathcal{H}_0}{P}(\mathbf{h})$ and all
$\sum_{\mathbf{h}\in\mathcal{H}_1}{P}(\mathbf{h})$
can also be seen in Figure~\ref{fig:BlindSrcSept_GRBM}
by comparing a GRBM-2-2 (a) with an two dimensional MoG having three isotropic components denoted by MoG-2-3 (d). Although the MoG-2-3 has one component fewer than the GRBM-2-2, their
probability distributions are almost the same.

\subsection{Natural image patches}
In contrast to random images, natural images have a common underlying structure
which could be used to code them more efficiently than with a pixel-wise representation.
\cite{OlshausenField-1996a} showed that sparse coding is such an efficient coding scheme
and that it is in addition a biological plausible model for the simple cells in
the primary visual cortex. 
\cite{BellSejnowski-1997} showed that the independent components provide a comparable representation for natural images. We now want to test empirically whether GRBMs generate such biological plausible results like sparse coding and ICA.

We used the $\mathtt{imlog}$ natural image Database of \cite{HaterenSchaaf-1998a} and
randomly sampled 70000 grey scale image patches with a size of $14\times14$ pixels. The
data was whitened using Zero-phase Component Analysis (ZCA), afterwards it was divided into 40000 training and 30000 testing image patches.
We followed the training recipes mentioned in
Section~\ref{sec:trainingReceipt}, since training a GRBM on natural image patches is not a trivial task.

\begin{figure}[ht] 
    \includegraphics[width=\linewidth]{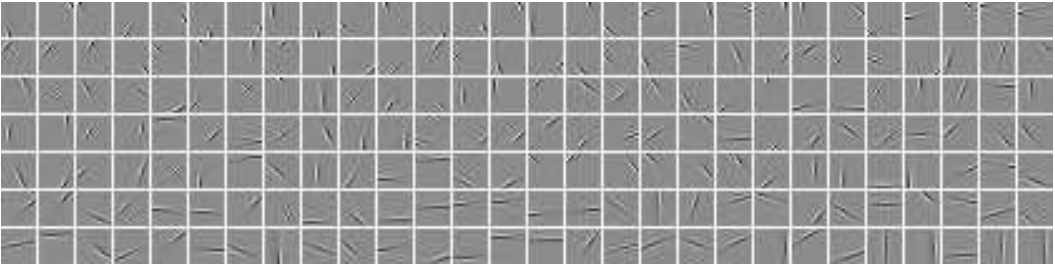}
\caption{Illustration of 196 learned filters of a GRBM-196-196. The plot has
been ordered from left to right and from top to bottom by the increasing
average activation level of the corresponding hidden units.} 
\label{fig:NIFiltersGBRBM}
\end{figure}

\begin{figure}[ht]
  \begin{center}
  \includegraphics[scale=0.4]{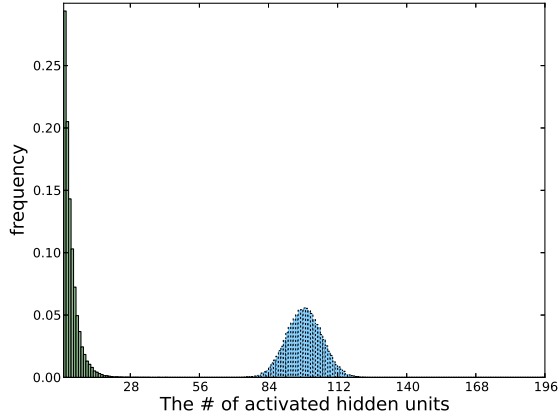}
  \end{center}
  \caption{The histogram of the number of activated hidden units per training sample. The histograms before and after training are plotted in blue (dotted) and in green (solid), respectively.}
  \label{fig:NIHiddenActivityHist}
\end{figure}

In Figure~\ref{fig:NIFiltersGBRBM}, we show the learned weight vectors namely features or filters, which can be
regarded as receptive fields of the hidden units. They are fairly similar to the
filters learned by ICA~\cite{BellSejnowski-1997}.  
Similar to the 2D experiment, we calculated the anchor and first order mixing coefficients, as shown in
Table~\ref{tab:NICoefficientAvgActiv}. The coefficients are much smaller compared to the anchor and first order coefficients of the GRBMs in the two dimensional case. 
However, they are still significantly large, considering that the total number of
components in this case is $2^{196}$. Similar to the two-dimensional
experiments, the more activated hidden units a state has, the less likely it will be reached,
which leads naturally to a sparse representation. To support
this statement, we plotted the histogram of the number of activated hidden units per training sample,
as shown in Figure~\ref{fig:NIHiddenActivityHist}.
\begin{table}[ht] 
  \caption{The mixing coefficients of GRBMs-196-196 per component (the Partition function was estimated using AIS).}
\label{tab:NICoefficientAvgActiv} 
\begin{center} 
\begin{tabular}{lccc} 
\multicolumn{1}{c}{} & $\sum\limits_{\mathbf{h}\in\mathcal{H}_0}{P}(\mathbf{h})$ 
    & $\sum\limits_{\mathbf{h}\in\mathcal{H}_1}{P}(\mathbf{h})$  & $\sum\limits_{\mathbf{h}\in{\mathcal{H} \backslash \lbrace \mathcal{H}_0 \cup \mathcal{H}_1 \rbrace }}{P}(\mathbf{h})$ \\
\hline \\ 
GRBM-196-196 & $0.04565$ & $0.00070$ & $0.95365$   \\ 
\end{tabular} 
\end{center} 
\end{table} 

\begin{figure}[ht]
  \begin{center}
  \includegraphics[scale=0.5]{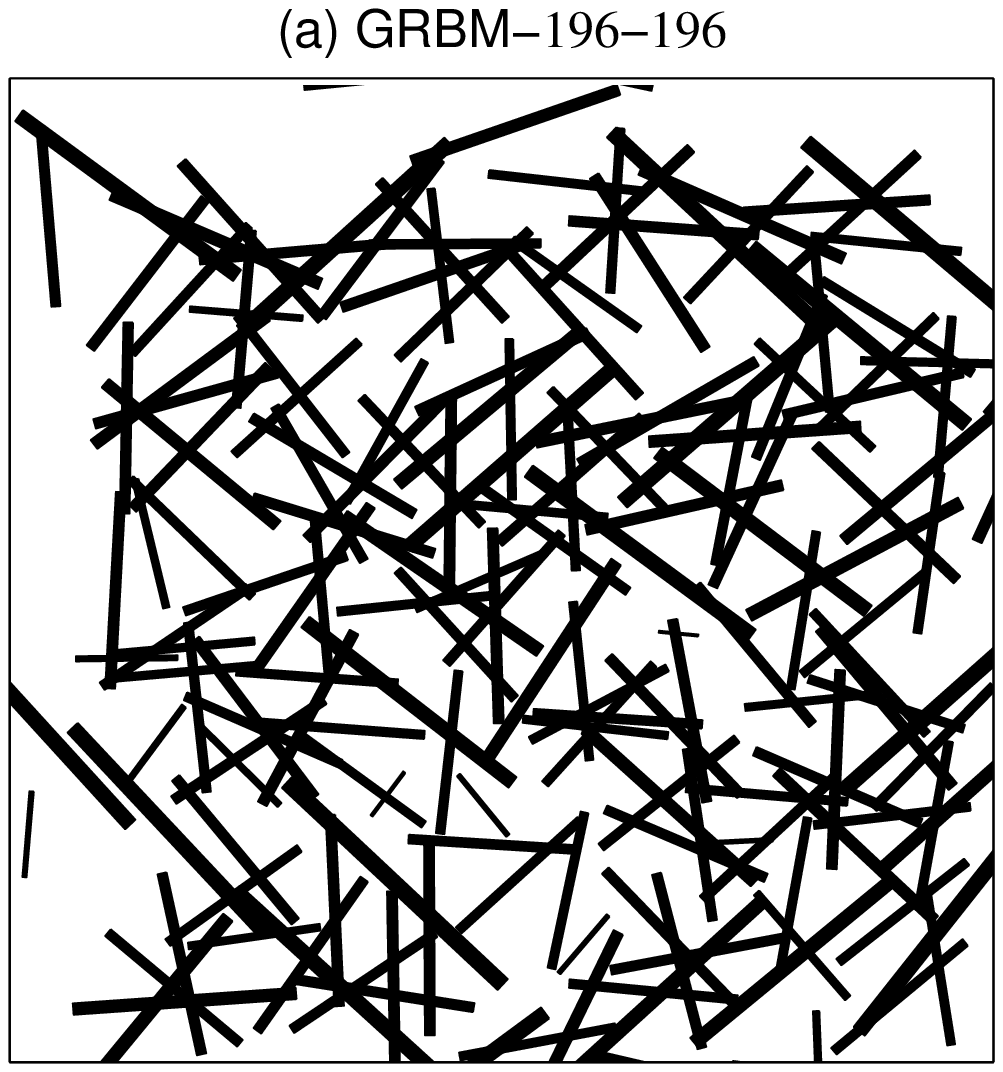}
  \includegraphics[scale=0.5]{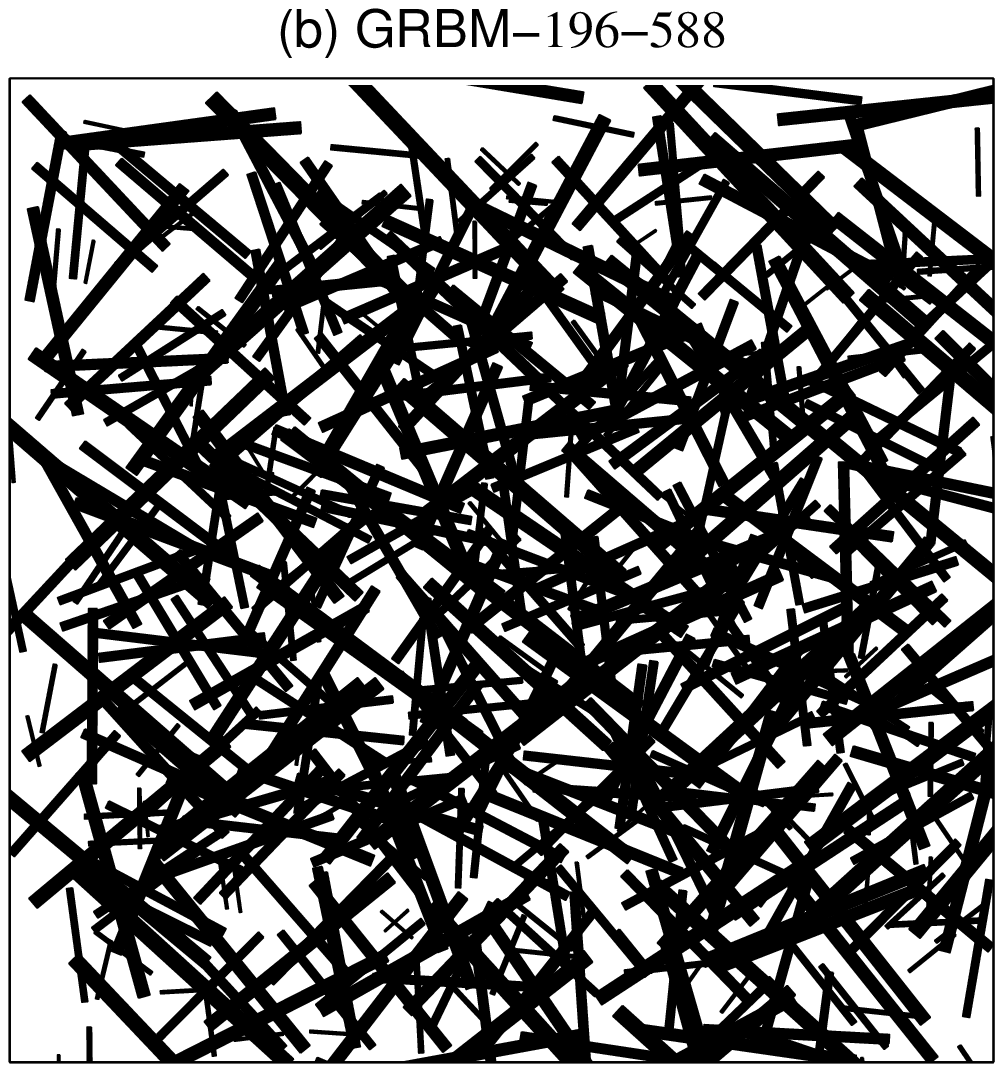}
  \end{center}
  \caption{The spatial layout and size of the filters, which are described by
  the position and size of the bars. Each bar denotes the center position and
  the orientation of a fitted Gabor function within $14\times14$ grid. The
  thickness and length of each bar are propotional to its spatial-frequency
  bandwidth.}
  \label{fig:NIFiltersSpatial}
\end{figure}

\begin{figure}[ht]
  \begin{center}
  \includegraphics[scale=0.5]{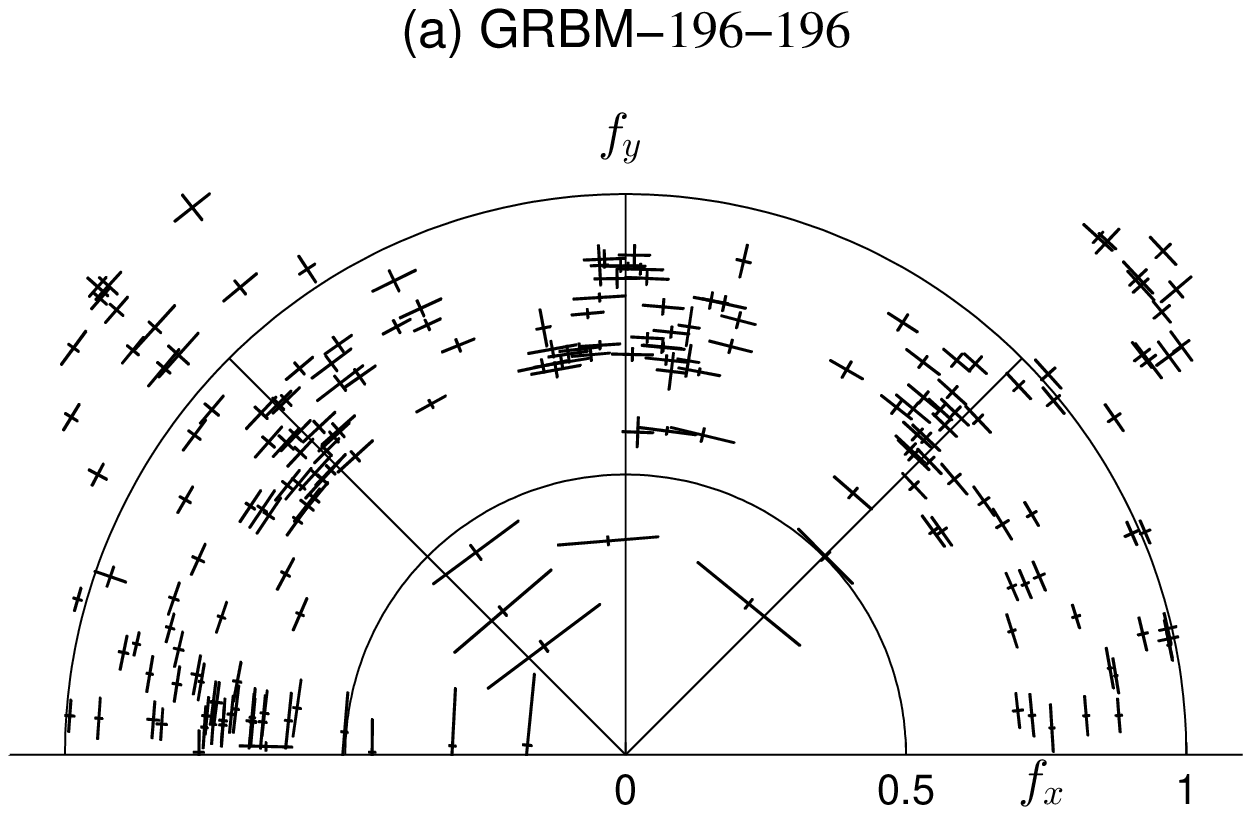}
  \includegraphics[scale=0.5]{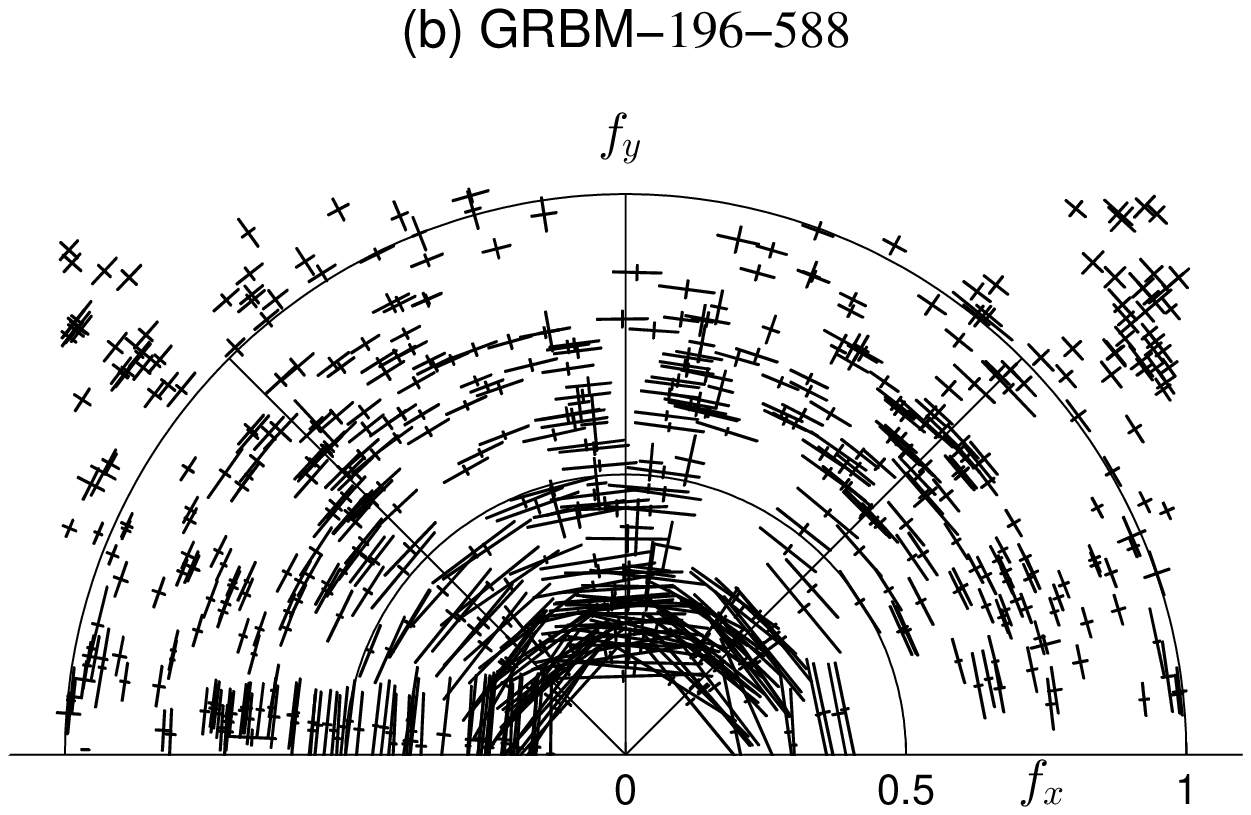}
  \end{center}
  \caption{A polar plot of frequency tuning and orientation of the
  learned filters. The crosshairs describe the selectivity of the filters, which is given by the $1/16$-bandwidth in
  spatial-frequency and orientation, \cite{LewickiOlshausen-1999a}. }
  \label{fig:NIFiltersPolar}
\end{figure}

We also examined the results of GRBMs in the over-complete case, i.e.
GRBM-196-588. There is no prominent difference of the filters compared to the complete case shown in Figure~\ref{fig:NIFiltersGBRBM}.
To further compare the filters in the complete and over-complete case, we estimated the spatial frequency, location and orientation for all filters in the spatial and frequency domains, see Figure~\ref{fig:NIFiltersSpatial} and Figure~\ref{fig:NIFiltersPolar} respectively. This is achieved by fitting a Gabor function of the form used by~\cite{LewickiOlshausen-1999a}. Note that the additional filters in the over-complete case increase the variety of spatial frequency, location and orientation.

\section{Successful Training of GRBMs on Natural Images}
\label{sec:trainingReceipt}

The training of GRBMs has been reported to be difficult~\cite{ Krizhevsky-2009a,ChoIlinEtAl-2011b}. 
Based on our analysis we are able to propose some recipes which should 
improve the success and speed of training GRBMs on natural image patches.
Some of them do not depend on the data distribution and should therefore improve the training in general.

\subsection{Preprocessing of the Data}
\label{sec:preprocessing}
The preprocessing of the data is important especially if the model is highly restricted like GRBMs.
Whitening is a common preprocessing step for natural images. It removes
the first and second order statistics from the data, so that it has zero mean
and unit variance in all directions. This allows training algorithms to focus
on higher order statistics like kurtosis, which is assumed to play an important
role in natural image representations~\cite{OlshausenField-1996a,
HyvarinenKarhunenEtAl-2001a}. 

The components of GRBMs are isotropic Gaussians, so that the model
would use several components for modeling covariances. But the whitened data has a spherical covariance matrix so that
the distribution can be modelled already fairly well by a single component. The other
components can then be used to model higher order statistics, so that we claim
that whitening is also an important preprocessing step for GRBMs.

\subsection{Parameter Initialization}

The initial choice of model parameters is important for
optimization process. Using prior knowledge about the optimization problem 
can help to derive an initialization, which can improve the speed and success of the training.

For GRBMs we know from the analysis above that the anchor component, which is placed at the visible bias, represents most of the whitened data.
Therefore it is reasonable in practice to set the visible bias to the data's mean.

Learning the right scaling is usually very slow since the weights and biases determine both the position and scaling of the components. 
In the final stage of training GRBMs on whitened natural images, the first 
components are scaled down extremely compared to
the anchor component. 
Therefore, it will usually speed up the training process if we initialize the parameters so that the first order scaling factors are already
very small. Considering equation~(\ref{eqn:probOfXExpandGauss}), we are able to
set a specific first order scaling factor by initializing the hidden bias to
\begin{equation}
	c_j = -\frac{||\mathbf{b}+\mathbf{w}_{* j}||^2 - ||\mathbf{b}||^2}{2\sigma^2} + \ln \tau_j ,
	\label{eqn:InitHiddenBias}
\end{equation}
so that the scaling is determined by $\tau_j$, which should ideally be
chosen close to the unknown final scaling factors. 
In practice, the choice of $0.01$ showed good performance in most cases.
The learning rate for the hidden bias can then be set much smaller than the 
learning rate for the weights.

According to~\cite{Bengio-2010}, the weights should be initialized to $w_{ij} \sim U\left(-\frac{\sqrt{6}}{\sqrt{N+M}}, \frac{\sqrt{6}}{\sqrt{N+M}} \right)$, where
$U(a, b)$ is the uniform distribution in the interval [a, b]. In our experience, this works better than the commonly used initialization to small Gaussian-distributed random values.

\subsection{Gradient Restriction and Choices of the Hyperparameters}
The choice of the hyper-parameters has an significant impact on the speed and success of training
GRBMs.  For successful training in an acceptable number of updates, the
learning rate needs to be sufficiently big.  Otherwise the learning process
becomes too slow or the algorithm converges to a local optimum where all
components are placed in the data's mean. But if the learning rate is chosen too
big, the gradient can easily diverge resulting in a number overflow of the
weights. This effect becomes even more crucial as the model dimensionality
increases, so that a GRBM with 196 visible and 1000 hidden units diverges
already for a learning rate of 0.001.

We therefore propose restricting the weight gradient column norms $\nabla w_{:j}$ to a meaningful size 
to prevent divergence. Since we know that the components are placed in the region of data,
there is no need for a weight norm to be bigger than twice the maximal data
norm. Consequently, this natural bound also holds for the gradient and can in
practice be chosen even smaller. It allows to choose big learning rates even
for very large models and therefore enables fast and stable training. 
In practice, one should restrict the norm of the update matrix rather than the gradient matrix to also restrict the effects of the momentum term and etc.

Since the components are placed on the data they are naturally restricted,
which makes the use of a weight decay useless or even counter productive since
we want the weights to grow up to a certain norm. Thus we do recommend not to use a weight decay regularization.

A momentum term adds a percentage of the old gradient to the current gradient
which leads to a more robust behavior especially for small batch-sizes. 
In the early stage of training the gradient usually varies a lot, a large momentum can therefore be used to prevent the weights from converging to zero. In the late stage however, it can also prevent convergence so that in practice a momentum of 0.9 that will be reduced to zero in the final stage of training is recommended. 

\subsection{Training Method}

Using the gradient
approximation, 
RBMs are usually trained as described in Section~\ref{sec:trainingAlgorithm}.
The quality of the approximation highly depends on the set of samples used for
estimating the model expectation, which should ideally be i.i.d.  But Gibbs
sampling usually has a low mixing rate, which means that the samples tend to
stay close to the previously presented samples. Therefore, a few steps of Gibbs
sampling commonly leads to a biased approximation of the gradient. In order to
increase the mixing rate \cite{Tieleman-2008a} suggested to use a persistent
Markov chain for drawing samples from the model distribution, which is referred
as persistent Contrastive Divergence (PCD). 
\cite{DesjardinsCourvilleEtAl-2010a} proposed to use parallel tempering (PT),
which selects samples from a persistent Markov chain with a different scaling
of the energy function. In particular, \cite{ChoIlinEtAl-2011b} analyzed PT
algorithm for training GRBMs and proposed a modified version of PT.

\begin{figure} 
  \begin{center}
	 \includegraphics[scale=0.5]{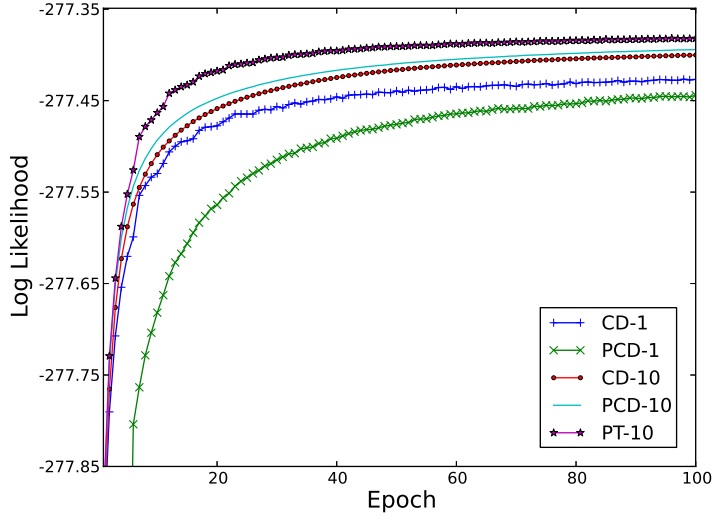}
  \end{center}
\caption{Evolution of the $\hat\ell$ of a GRBM 196-16 on the whitened natural
image dataset for CD, PCD using a $k$ of $1$, $10$ each and PT with 10
temperatures. The learning curves are the average over 40 trials. The learning
rate was 0.1, an initial momentum term of 0.9 was multiplied with 0.9 after
each fifth epoch, the gradient was restricted to one hundredth of the maximal
data norm (0.48), no weight decay was used. } 
\label{fig:CDvsPT}
\end{figure}

In our experiments all methods above lead to meaningful features and comparable
$\hat\ell$, but differ in convergence speed as shown in
Figure~\ref{fig:CDvsPT}. As for PT, we used original algorithm
\cite{DesjardinsCourvilleEtAl-2010a} together with weight restrictions and
temperatures from 0.1 to 1 with step-size 0.1. Although, PT has a better
performance than CD, it has also a much higher computational cost as shown in
Table~\ref{tab:MethodCPU}. 

\begin{table}
\centering
      \begin{tabular}{|c|c|}\hline
      Method & Time per epoch in s \\ \hline
					      CD-1  & 2.1190 \\ 
			PCD-1   & 2.1348 \\ 
			CD-10 & 10.8052  \\ 
			PCD-10  & 10.8303 \\ 
			PT-10  & 21.4855 \\
			\hline
     \end{tabular}
     \caption{Comparison of the CPU time for training a GRBM with different methods.} 
     \label{tab:MethodCPU}
\end{table}

\section{Discussion}
The difficulties of using GRBMs for modeling natural images have been reported
by several authors~\cite{Krizhevsky-2009a, BengioLamblinEtAl-2006a} and
various modifications have been proposed to address this problem.

\cite{RanzatoHinton-2010a} analyzed the problem from the view of generative
models and argued that the failure of GRBMs is due to the model's focus
on predicting the mean intensity of each pixel rather than the dependence between pixels. To model the
covariance matrices at the same time, they proposed the
mean-covariance RBM (mcRBM). In addition to the conventional hidden units
$\mathbf{h}^m$, there is a group of hidden
units $\mathbf{h}^c$ dedicated to model the covariance between the visible units. 
From the view of density models, mcRBMs can be regarded as improved GRBMs such that the
additional hidden units are used to depict the covariances.
The conditional probabilities of mcRBM are given by
\begin{equation}
  P(\mathbf{X}|\mathbf{h}^m, \mathbf{h}^c) = 
  \mathcal{N}\left( \mathbf{X}; \boldsymbol\Sigma
   \,\mathbf{W}\,\mathbf{h}^m  
   , \boldsymbol\Sigma\right),
  \label{eqn:ProbmcRBM}
\end{equation}
where $\boldsymbol\Sigma = \left( \mathbf{C}\,diag(\mathbf{P}\mathbf{h}^c)\,\mathbf{C}^T \right)^{-1}$ 
\cite{RanzatoHinton-2010a}.
By comparing (\ref{eqn:ProbmcRBM}) and (\ref{eqn:probOfXHGauss}), it can be seen that the components of mcRBM can have a covariance matrix that is not restricted to be diagonal as it is the case for GRBMs.

From the view of generative models another explanation for the failure of GRBMs
is provided by~\cite{CourvilleBergstraEtAl-2011a}. Although they agree with
the poor ability of GRBMs in modeling covariances,
~\cite{CourvilleBergstraEtAl-2011a} argue that the deficiency is due to
the binary nature of the hidden units. In order to overcome this limitation,
they developed the spike-and-slab RBM (ssRBM), which splits each binary hidden
unit into a binary spike variable $h_j$ and a real valued slab variable $s_j$.
The conditional probability of visible units is given by
\begin{equation}
  P(\mathbf{X}|\mathbf{s},\mathbf{h},||\mathbf{X}||^2<R) = 
  \frac{1}{B} \mathcal{N}\left(\mathbf{X}; 
  \boldsymbol\Lambda^{-1} \sum_{j=1}^N{\mathbf{w}_{* j}s_jh_j}, \boldsymbol\Lambda^{-1} \right),
\label{eqn:ssRBM}
\end{equation}
where $\boldsymbol\Lambda$ is a diagonal matrix and $B$ is determined by
integrating the Gaussian \linebreak      
    $\mathcal{N}(\mathbf{X}; 
  \boldsymbol\Lambda^{-1} \sum_{j=1}^N{\mathbf{w}_{* j}s_jh_j}, \boldsymbol\Lambda^{-1} )$       
over the ball $||\mathbf{X}||^2<R$ ~\cite{CourvilleBergstraEtAl-2011a}. In
contrast to the conditional probability  of GRBMs (\ref{eqn:probOfXHGauss}),
$\mathbf{w}_{* j}$ in (\ref{eqn:ssRBM}) is scaled by the continuous variable
$s_j$, which implies that the components can be shifted along their weight
vectors.

We have shown that GRBMs are capable of modeling natural image patches and that
the reported failures are due to the training procedure.
\cite{LeeEkanadhamEtAl-2007a} showed also that GRBMs could learn meaningful
filters by using a sparse penalty. But this penalty changes the objective
function and introduced a new hyper-parameter. 

\cite{ChoIlinEtAl-2011b} addressed these training difficulties, by
proposing a modification of PT and an adaptive learning rate.  However, we
claim that the reported difficulties of training GRBMs with PT are due to the
mentioned gradient divergence problem. With gradient restriction we were able
to overcome the problem and train GRBMs with normal PT successfully. 

\section{Conclusion}
In this paper, we provide a theoretical analysis of GRBM and showed that its
product of experts formulation can be rewritten as a constrained mixture of
Gaussians. This representation gives a much better insight into the
capabilities and limitations of the model.  
We use two-dimensional blind source separation task as a toy problem to demonstrate how GRBMs
model the data distribution. In our experiments, GRBMs were capable of learning
meaningful features both in the toy problem and in modeling natural images.

In both cases, the results are comparable to that of ICA.
But in contrast to ICA the features are not restricted to be orthogonal
and can form an over-complete representation.
However, the success of training GRBMs highly depends on the training setup,
for which we proposed several recipes based on the theoretical analysis. Some
of them can be further generalized to other datasets or directly applied like
the gradient restriction. 

In our experience, maximizing the $\hat\ell$ does not imply good features and vice versa.
Prior knowledge about the data distribution will be
beneficial in the modeling process. For instance, our recipes are based on the
prior knowledge of the natural image statistics, which is center peaked and has
heavy tails. It will be an interesting topic to integrate
prior knowledge of the data distribution into the model rather than starting
modeling from scratch.

Considering the simplicity and easiness of training with our proposed recipe, we believe that GRBMs provide a possible way for modeling natural images. 
Since GRBMs are usually used as first layer in deep belief networks, the successful training of GRBMs should therefore improve the performance of the whole network.

\end{document}